\documentclass[ited]{informs4}
\usepackage{eqndefns-left}
\RequirePackage{newtxtext}
\RequirePackage{newtxmath}
\RequirePackage{endnotes}

\OneAndAHalfSpacedXII

\usepackage{etoolbox}

\AtBeginEnvironment{algorithmic}{%
  \small
  \setlength{\baselineskip}{0.7\baselineskip}% tweak factor as desired
}

\usepackage{caption}
\usepackage{booktabs}
\usepackage{tabularx}
\usepackage{hyperref}
\usepackage{multirow}
\usepackage{algorithm}
\usepackage{algpseudocode}
\usepackage[section]{placeins}
\usepackage{subcaption}

\usepackage{natbib}

\bibpunct[, ]{(}{)}{,}{a}{}{,}%
\def\bibfont{\small}%
\EquationsNumberedThrough    % Default: (1), (2), ...

\TheoremsNumberedThrough     % Preferred (Theorem 1, Lemma 1, Theorem 2)
\ECRepeatTheorems

\begin{document}
%%%%%%%%%%%%%%%%

\RUNAUTHOR{De Kerpel L., Thuy A., and Benoit D.F.}

% (shortened) title
\RUNTITLE{Bandits for Educational Recommender Systems}

% Full title
\TITLE{A Bandit-Based Approach to Educational Recommender Systems: Contextual Thompson Sampling for Learner Skill Gain Optimization}

\ARTICLEAUTHORS{%
\AUTHOR{Lukas De Kerpel}
\AFF{Faculty of Economics and Business Administration, Ghent University;\\
 FlandersMake@UGent, corelab CVAMO, Tweekerkenstraat 2, 9000, Ghent, Belgium,\\
 \EMAIL{lukas.dekerpel@ugent.be}}

\AUTHOR{Arthur Thuy}
\AFF{Faculty of Economics and Business Administration, Ghent University;\\
 FlandersMake@UGent, corelab CVAMO, Tweekerkenstraat 2, 9000, Ghent, Belgium,\\
 \EMAIL{arthur.thuy@ugent.be}}

\AUTHOR{Dries F. Benoit}
\AFF{Faculty of Economics and Business Administration, Ghent University;\\
 FlandersMake@UGent, corelab CVAMO, Tweekerkenstraat 2, 9000, Ghent, Belgium,\\
 \EMAIL{dries.benoit@ugent.be}}
}

\ABSTRACT{
In recent years, instructional practices in Operations Research (OR), Management Science (MS), and Analytics have increasingly shifted toward digital environments, where large and diverse groups of learners make it difficult to provide practice that adapts to individual needs. 
This paper introduces a method that generates personalized sequences of exercises by selecting, at each step, the exercise most likely to advance a learner’s understanding of a targeted skill. 
The method uses information about the learner and their past performance to guide these choices, and learning progress is measured as the change in estimated skill level before and after each exercise. 
Using data from an online mathematics tutoring platform, we find that the approach recommends exercises associated with greater skill improvement and adapts effectively to differences across learners. 
From an instructional perspective, the framework enables personalized practice at scale, highlights exercises with consistently strong learning value, and helps instructors identify learners who may benefit from additional support.
}

% Keywords

%\HISTORY{Received: Month DD, YYYY; Accepted: Month DD, YYYY; Published Online: Month DD, YYYY}

\maketitle

%%%%%%%%%%%%%%%%%%%%%%%%%%%%%%%%%%%%%%%%%%%%%%%%%%%%%%%%%%%%%%%%%%%%

\section{Introduction}\label{sec:Intro}

Over recent years, active learning has been increasingly adopted in OR, MS, and Analytics education, transforming how students engage with learning materials, receive feedback, and develop conceptual understanding \citep{10.1287/ited.2021.0246,10.1287/ited.2023.0020,doi:10.1287/ited.2021.0243}. 
One way active learning strategies have been implemented is through Massive Open Online Courses (MOOCs) and other digital learning environments that provide online access to instructional materials, practice activities, and assessments. 
Yet despite this digital shift in instructional practice, many courses continue to rely on standardized learning paths in which all learners progress through the same predetermined sequence of exercises, offering limited opportunity for personalized learning experiences. 
The need for adaptive sequencing becomes even more difficult to address in MOOCs, where the sheer number of learners severely constrains instructors’ ability to provide individualized learning trajectories. 
Moreover, students in OR/MS/Analytics courses enter with widely differing levels of quantitative skills, yet scalable mechanisms for adapting practice to these varied skill levels remain limited.

Educational recommender systems (ERS) offer a promising approach for supporting active learning by tailoring practice opportunities to learners’ evolving needs and guiding them along personalized learning paths. 
Within such systems, personalized recommendations function as a form of individualized feedback through scaffolding.
Scaffolding is traditionally understood as an instructional practice in which a teacher provides structured support and gradually removes guidance as learners develop greater competence \citep{van_de_pol_scaffolding_2010}.
It involves sequencing tasks so that each exercise offers an appropriate level of challenge relative to the learner’s current understanding. 
In our setting, the ERS operationalizes this principle by continuously updating its estimates of a learner’s skill state after each interaction and selecting subsequent exercises whose difficulty is aligned with that learner’s evolving proficiency. 
In doing so, the ERS provides individualized feedback in the form of targeted guidance about what to practice next.

In the ERS literature, the most commonly used approach for generating recommendations has been collaborative filtering (CF) techniques \citep{khanal_systematic_2020}. 
CF methods identify patterns in interaction logs to recommend relevant exercises to users, typically through the detection of similarity either between users (UserCF) or between exercises (ItemCF). 
Nevertheless, CF methods face important limitations in educational contexts. 
First, CF methods are not inherently personalized: recommendations are derived from aggregated behavioral patterns rather than the learner's unique profile. 
User-based CF, for example, assumes that learners with similar past interactions will benefit from similar exercises, overlooking individual heterogeneity in learning needs or cognitive skill levels.
Second, learners’ preferences and knowledge states evolve over time, yet CF approaches typically rely on static similarity measures.
This makes CF ill-suited to capture temporal dependencies or to adjust recommendations as learners progress. 
Third, without an explicit mechanism for exploration, CF also tends to reinforce historically popular exercises, limiting opportunities to identify exercises better aligned with a learner’s evolving profile.

Contextual bandit algorithms, by contrast, are inherently adaptive and address these shortcomings directly. 
They frame recommendation as a sequential decision problem in which the effectiveness of an exercise is uncertain \emph{ex ante} and heterogeneous across learners, skills, and time. 
In a bandit setting, each recommendation step selects an exercise for a particular learner. 
The ``context'' is the information available at decision time about the learner and the candidate exercise (e.g., recent performance, affective state, topic, difficulty), and the resulting ``reward'' is the observed learning benefit after engagement, here operationalized as the change in estimated skill gain. 
The exploration–exploitation mechanism at the heart of bandit methods enables deliberate trialing of uncertain but potentially valuable exercises (\textit{exploration}) while selecting those exercises expected to generate the greatest learning gains (\textit{exploitation}). 
In this way, bandit algorithms can contribute to the construction of adaptive learning paths that respond to evolving learner needs. 
The literature study by \citet{da_silva_systematic_2023} highlights that, despite their promise, bandit-based methods remain underexplored in ERS, underscoring a key avenue for future research.

This study addresses the identified literature gap by proposing a bandit-based framework for ERS and, to the best of our knowledge, is the first empirical evaluation of Thompson Sampling for educational recommendation. 
Thompson Sampling (TS) is a Bayesian posterior sampling algorithm \citep{thompson1} with strong theoretical regret guarantees \citep{agrawal1,agrawal2} and robust empirical performance across sequential decision-making tasks \citep{ferreira1,aramayo1}. 
At each decision step, TS maintains a posterior distribution over the expected utility of the available exercises and selects the next exercise by sampling from these distributions.
In doing so, exercises are chosen in proportion to their probability of being optimal, which provides a principled mechanism for balancing \textit{exploration} of uncertain exercises with \textit{exploitation} of those exercises expected to yield high utility.

The proposed framework applies Linear Thompson Sampling (LinTS) as the bandit strategy for educational recommendation. 
LinTS specifies a separate linear model for each available exercise, where the expected reward is expressed as a linear function of learner features. 
This formulation enables recommendations to adapt to evolving learner profiles and knowledge states. 
For benchmarking, we also consider standard TS, which operates in a non-contextual setting, alongside conventional CF approaches. 
This comparative design enables a systematic evaluation of the contribution of contextual modeling to sequential recommendation in ERS.

The framework employs a reward signal based on learner skill gain, defined as the learner’s improvement in estimated knowledge state of a particular cognitive skill as computed by a Bayesian Knowledge Tracing (BKT) model \citep{corbett_knowledge_1994}. 
BKT models skill acquisition as a latent probabilistic process, updating a learner’s knowledge state with each interaction based on observed responses, thereby providing a dynamic estimate of learning progress. 
The use of skill gain as reward contrasts with the predominant evaluation metrics in recommender studies, which rely on the correctness of recommended exercises \citep{lan_contextual_2016}, ratings of recommended exercises \citep{nafea_recommendation_2019}, or user satisfaction levels \citep{tarus_hybrid_2017}.
Correctness may lead to inflated performance estimates, as systems can achieve high accuracy by recommending exercises the learner is already able to solve, without necessarily promoting new learning. 
Similarly, ratings and satisfaction scores capture subjective perceptions of difficulty or enjoyment but may diverge from actual knowledge acquisition. 
These metrics are therefore limited as proxies for genuine learning progress, as they do not fully capture the cognitive development of a learner. 
Accordingly, skill gain is adopted as the reward signal, as it more directly aligns the optimization process with the pedagogical objective of competence development.

Our experiment is performed on the ASSISTments dataset \citep{patikorn_assistments_2020}, which provides interaction data from an online secondary-level tutoring system \citep{heffernan_assistments_2014}. 
The reward signal is defined as the change in estimated skill between two consecutive interactions, based on the BKT proficiency measures included in the dataset. 
Results show that LinTS achieves the highest performance, yielding a 15.2\% improvement in average skill gain over the non-contextual TS baseline, as well as 16.5\% and 20.7\% improvements over CF baselines. 
This underscores the effectiveness of contextual modeling for adaptive educational recommendation.

The remainder of this paper is organized as follows.
Section~\ref{sec:RelatedWork} reviews related work on ERS and bandit methods.
Section~\ref{sec:Method} formalizes the problem setting and details the baselines (UserCF, ItemCF) and the bandit policies (TS, LinTS).
Section~\ref{sec:exp_settings} describes the experimental setup, including dataset description, data preprocessing, data splitting, and validation strategy. 
Section~\ref{sec:Results} reports empirical results, while Section~\ref{sec:Discussion} discusses practical takeaways for OR/MS/Analytics instructors.
Section~\ref{sec:Conclusion} ends with a conclusion and directions for future work.

\section{Related work}\label{sec:RelatedWork}

Research on active learning in OR/MS/Analytics education has primarily focused on improving in-class teaching practices. 
For example, \citet{10.1287/ited.2023.0020} examined efficient approaches for assessing higher-order thinking skills in an undergraduate business analytics course.
Similarly, \citet{doi:10.1287/ited.2021.0243} introduced a puzzle-based learning method to support the development of optimization skills in high school students, while \citet{10.1287/ited.2021.0246} designed hands-on, constructivist exercises to enhance learning in an introductory probability and statistics course. 
All these interventions remain largely classroom-based. 
Importantly, none of these studies consider how active learning might be supported through personalized learning paths in digital environments, a domain that is becoming increasingly relevant as OR/MS/Analytics courses move online.

Several approaches have been proposed to optimize sequences of exercises in ERS, each aiming to guide learners through personalized educational trajectories. 
CF remains the most dominant approach in ERS, with five out of the sixteen studies analyzed in the literature review by \citet{da_silva_systematic_2023} employing either pure CF methods or hybrid strategies that incorporate CF components. 
CF approaches are typically classified into two main variants: user-based and item-based filtering. 
The user-based approach infers preferences by identifying patterns of similarity among users, generating recommendations based on exercises previously interacted with by users exhibiting comparable behavioral profiles. 
Conversely, the item-based variant focuses on relationships between exercises, recommending exercises that are similar in content or usage patterns to those a given learner has already engaged with \citep{da_silva_systematic_2023}.
Among CF approaches, user-based variants are more commonly adopted in the recommendation process, reflecting the growing pedagogical emphasis on student-centered learning \citep{krahenbuhl_student_2016}. 
In the reviewed studies, various similarity metrics have been applied to operationalize these patterns, including cosine similarity \citep{wu_fuzzy_2015, tarus_hybrid_2017, huang_score_2019}, Euclidean distance \citep{sergis_learning_2016}, and Pearson correlation coefficient \citep{nafea_recommendation_2019}, each offering different computational strategies for capturing relational proximity within interaction data.

While CF has shown effectiveness in static recommendation scenarios, its reliance on historical similarity patterns constrains its pedagogical applicability. 
Recommendations are derived from aggregated interaction behavior rather than being informed by the learner’s evolving knowledge state, and the absence of an explicit exploration mechanism leads CF to reinforce familiar or popular exercises instead of identifying exercises that may more effectively promote sustained learning progress. 
These limitations create the need for methods that explicitly model sequential decision-making.

Contextual bandit algorithms provide such a framework by conditioning recommendations on learner-specific features (personalization) and dynamically balancing the trade-off between exploiting known effective exercises and exploring uncertain but potentially more beneficial ones. 
As highlighted in the systematic literature review by \citet{da_silva_systematic_2023}, despite the prominence of bandit algorithms in general recommender system research, such methods remain underexplored in educational contexts, revealing a notable mismatch between the two domains and underscoring the need for further research.

Most existing applications of multi-armed bandit algorithms within the educational domain are situated in the context of educational games, where their capacity for adaptive, sequential decision-making aligns well with the dynamic and interactive nature of game-based learning environments. 
\citet{liu_trading_2014} apply a UCB-Explore strategy, a non-contextual bandit algorithm based on the Upper Confidence Bound (UCB) algorithm, in a physics-based educational game involving different ways of displaying number lines. 
\citet{clement_multi-armed_2015} employ the exponential-weight algorithm called EXP4, a context-free bandit strategy that incorporates expert knowledge to narrow the set of possible exercises when training a bandit policy. 
This method is applied within an educational game for 7- to 8-year-old schoolchildren, designed to support the development of numerical decomposition skills in the context of manipulating money. 
These studies are limited in two respects.
First, they target narrow, game-based case studies with bespoke mechanics and short interaction horizons; as a result, their findings may not transfer to broader settings such as MOOCs or large-scale tutoring platforms that involve heterogeneous content, longer trajectories, and diverse learner populations.
Second, both methods are \emph{context-free} and therefore cannot condition recommendations on learner features, precluding meaningful personalization.

To date, only two studies have applied contextual bandits within the ERS domain. 
The first study by \citet{lan_contextual_2016} employs a contextual Linear UCB algorithm with context vectors representing latent concept knowledge profiles inferred from learners’ interaction histories, in a college-level physics setting. 
The second study by \citet{intayoad_reinforcement_2020} incorporates past student behaviors and current learner state into a correlation analysis to pre-select candidate exercises. 
As such, we argue that this approach cannot be considered a true contextual bandit model, as contextual information is confined to the pre-filtering stage rather than being embedded within the bandit algorithm itself. 
The recommendation policy relies on a non-contextual $\epsilon$-greedy strategy, with reward defined as whether the learner clicked on the suggested exercise. 
While these approaches provide valuable empirical insights, the use of correctness on the next exercise in the first study and click-based feedback in the second as optimization criteria may be limiting for ERS. 
Such measures primarily capture short-term task performance or engagement and may fail to account for sustained learning gains or the incremental development of a learner’s knowledge state, which are particularly critical for constructing effective learning paths. 
In particular, the first study by \citet{lan_contextual_2016} incorporated knowledge estimates as input features to guide recommendations. 
However, we argue that it would be more appropriate to optimize directly for improvements in these knowledge states, thereby aligning system objectives with sustained learning progress rather than immediate task performance.

To date, no studies on multi-armed bandits in ERS have incorporated TS \citep{thompson1}, despite its well-documented effectiveness in general recommender system research \citep{ferreira1,aramayo1, 10.1145/3771931} and its empirically demonstrated superiority over frequentist strategies such as UCB \citep{chapelle_empirical_2011}. 
Its probabilistic exercise-selection mechanism facilitates a principled balance between exploration and exploitation while inherently accounting for uncertainty in reward estimation. 
The absence of TS in prior ERS research therefore constitutes a notable literature gap, which the present study seeks to address. 
In this study, we implement LinTS, a contextual bandit algorithm that models each exercise as a linear function of learner features, and evaluate it alongside the non-contextual TS baseline. 
Unlike prior approaches, we define the reward as skill gain, thereby directly optimizing for improvements in learners’ knowledge states rather than correctness on the next exercise.

\section{Methodology} \label{sec:Method}

This section describes the algorithms and decision-making framework implemented in our ERS experiment. 
We first introduce the multi-armed bandit (MAB) problem and its contextual variant. 
We then detail the CF baselines (UserCF and ItemCF) and the two bandit-based methods TS and LinTS.

\subsection{Multi-Armed Bandits in Educational Recommendation}

In the MAB framework, a recommendation session unfolds over a sequence of \(T\) discrete interaction rounds.
At each round \(t \in \{1, \dots, T\}\), the ERS selects an exercise \(a_t\) from a finite set of available exercises \(\mathcal{A}\). 
Each exercise corresponds to a distinct learning activity, such as a practice exercise, instructional video, or interactive simulation. 
Once the learner engages with the recommended exercise, the system observes a reward \(r_{t,a_t}\), drawn from an unknown probability distribution specific to that exercise.

In this work, the reward is defined as the skill gain associated with the specific cognitive skill $s$ targeted by the recommended exercise $a_t$.
Let \(K^{(s)}_{t-1}\) denote the learner’s estimated mastery of skill \(s\) immediately prior to the interaction, and \(K^{(s)}_{t}\) the mastery estimate immediately after, both estimated using a BKT model \citep{corbett_knowledge_1994}.
The reward is then computed as
\begin{equation}
r_{t,a_t} = K^{(s)}_{t} - K^{(s)}_{t-1}
\end{equation}
ensuring that the gain measure reflects changes in the learner’s knowledge state for the relevant skill only, rather than general performance or unrelated knowledge.
This continuous-valued reward formulation aligns the optimization objective with long-term pedagogical effectiveness, as it directly quantifies the incremental learning benefit rather than short-term correctness.

The goal of the ERS is to choose exercises \(\{a_t\}_{t=1}^T\) such that the sum of the learner’s realized rewards is maximized:
\begin{equation}
\max_{a_1, \dots, a_T} \ \sum_{t=1}^T r_{t,a_t}
\end{equation}
subject to the constraint that the expected reward associated with a particular exercise is initially unknown and must be inferred through interaction.
This creates a fundamental trade-off between \emph{exploration}---selecting exercises to improve knowledge about their effectiveness---and \emph{exploitation}---selecting exercises currently believed to yield the highest educational benefit.

The contextual extension of MABs (CMAB) allows the ERS to incorporate side information about the learner and the exercise.
At round $t$, the system observes a context vector $\mathbf{x}_t \in \mathbb{R}^d$ that includes learner features such as demographic attributes, historical performance or emotional state.
The expected reward for an exercise is then modeled as:
\begin{equation}
\mu_a(\mathbf{x}_t) = f(\mathbf{x}_t, a),
\end{equation}
where $f$ is an unknown reward function.
By leveraging context, CMAB algorithms can tailor recommendations to individual learners, potentially improving personalization and long-term learning outcomes.

\begin{figure}[htbp]
  \centering
  \includegraphics[width=0.8\linewidth]{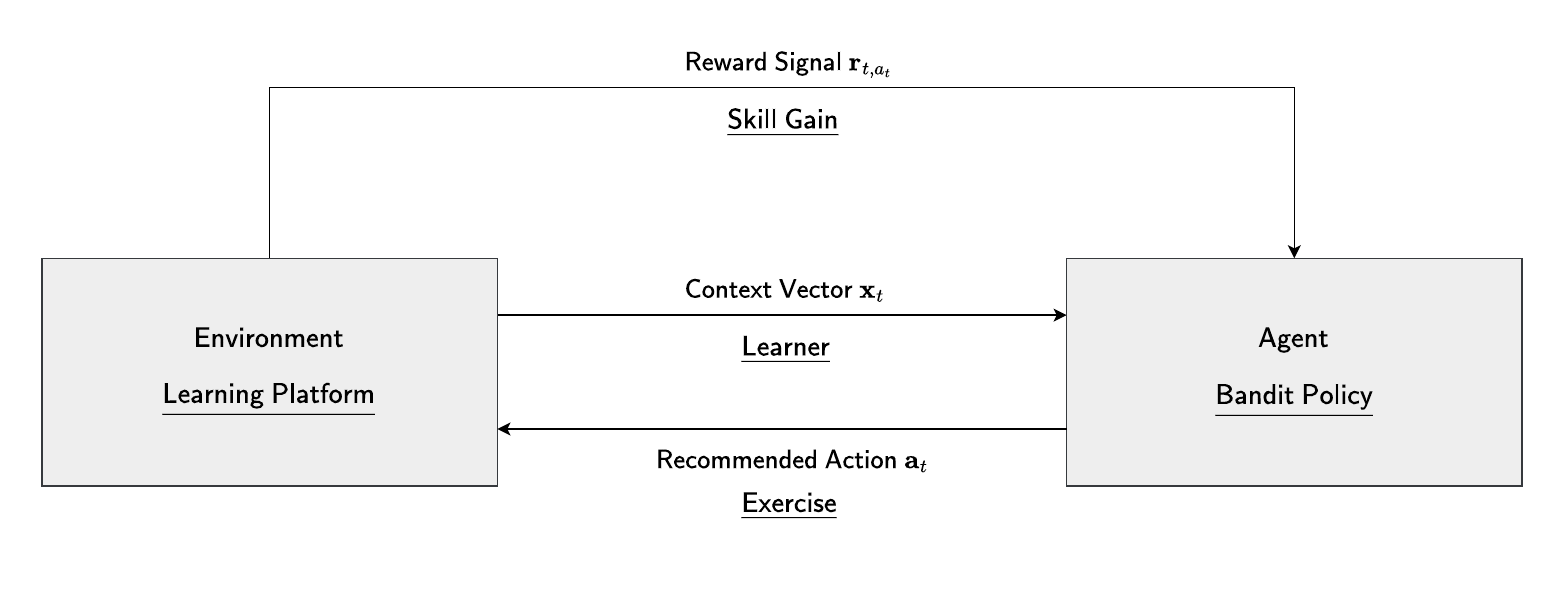}
  \caption{Bandit feedback process in an ERS.
   The environment (learning platform) emits context $\mathbf{x}_t$; the agent (bandit policy) recommends an exercise $a_t$; after the learner engages, the environment returns reward $r_{t,a_t}$ (skill gain).
   The resulting tuples $(\mathbf{x}_t,a_t,r_{t,a_t})$ support online learning and evaluation.}
  \label{fig:bandit_flow_ers}
\end{figure}

Figure~\ref{fig:bandit_flow_ers} summarizes the contextual bandit interaction in our ERS.
At time $t$, the learning platform (environment) provides a context vector $\mathbf{x}_t$ describing the current learner state.
The bandit policy (agent) selects an exercise $a_t$, which is delivered to the learner.
After the interaction, the platform returns a reward $r_{t,a_t}$ and the tuple $(\mathbf{x}_t,a_t,r_{t,a_t})$ is logged to update the policy.

\subsection{Collaborative filtering baselines}

\subsubsection{UserCF}
The user-based CF baseline estimates the expected effectiveness of a candidate exercise for a target learner by leveraging similarity in historical interaction profiles across the entire learner population.
Let $U$ denote the set of all learners, and let $R_{u,a}$ represent the observed skill-gain reward for learner $u$ on exercise $a$.
For a given target learner $u$, a similarity score $\mathrm{sim}(u,u')$ is computed with every other learner $u' \in U - \{u\}$ based on the cosine distance between their respective interaction vectors in the learner–exercise space.
Cosine-based similarity is the most common choice in distance-based CF, although alternative measures such as the Pearson correlation or the dot product are also used \citep{10.1145/3022227.3022292}.

The predicted effectiveness of an exercise $a$ is then obtained as a similarity-weighted average of the recorded rewards from all other learners:
\begin{equation}
\hat{R}_{u,a} = \frac{\sum_{u' \in U - \{u\}} \mathrm{sim}(u,u') \cdot R_{u',a}}{\sum_{u' \in U - \{u\}} |\mathrm{sim}(u,u')|}.
\end{equation}
The system then selects the exercise $a^{\star}$ with the maximal predicted value $\hat{R}_{u,a}$.
The UserCF algorithm is formally defined in Algorithm~\ref{alg:usercf}.

\begin{algorithm}
\caption{User-Based Collaborative Filtering} \label{alg:usercf}
\begin{algorithmic}[1]
\State \textbf{Input:} Target learner $u$, candidate exercise set $\mathcal{A}$, interaction matrix $R$ (skill gain)
\State Compute $\mathrm{sim}(u,u')$ for all $u' \in U - \{u\}$
\For{each $a \in \mathcal{A}$}
  \State $\hat{R}_{u,a} \gets \dfrac{\sum_{u' \in U - \{u\}} \mathrm{sim}(u,u')\,R_{u',a}}{\sum_{u' \in U - \{u\}} |\mathrm{sim}(u,u')|}$
\EndFor
\State \textbf{Return} $a^{\star} \gets \arg\max_{a \in \mathcal{A}} \hat{R}_{u,a}$
\end{algorithmic}
\end{algorithm}

\subsubsection{ItemCF}
The item-based CF baseline instead exploits similarities between exercises, as derived from their historical usage patterns across learners.
As in the user-based variant, $R_{u,a}$ stores the skill-gain reward for learner $u$ on exercise $a$.
For a target learner $u$ and a candidate exercise $a$, the cosine similarity $\mathrm{sim}(a,a')$ is computed with every other exercise $a' \in \mathcal{A} - \{a\}$.
The predicted skill-gain reward for $a$ is then given by the similarity-weighted average of the learner's own past rewards on other exercises:
\begin{equation}
\hat{R}_{u,a} = \frac{\sum_{a' \in \mathcal{A} - \{a\}} \mathrm{sim}(a,a') \cdot R_{u,a'}}{\sum_{a' \in \mathcal{A} - \{a\}} |\mathrm{sim}(a,a')|}.
\end{equation}
The algorithm then recommends the exercise $a^{\star}$ with the highest predicted value.
The ItemCF algorithm is formally defined in Algorithm~\ref{alg:itemcf}.

\begin{algorithm}
\caption{Item-Based Collaborative Filtering} \label{alg:itemcf}
\begin{algorithmic}[1]
\State \textbf{Input:} Target learner $u$, candidate exercise set $\mathcal{A}$, interaction matrix $R$ (skill gain)
\State Compute $\mathrm{sim}(a,a')$ for all $a' \in \mathcal{A} - \{a\}$
\For{each $a \in \mathcal{A}$}
  \State $\hat{R}_{u,a} \gets \dfrac{\sum_{a' \in \mathcal{A} - \{a\}} \mathrm{sim}(a,a')\,R_{u,a'}}{\sum_{a' \in \mathcal{A} - \{a\}} |\mathrm{sim}(a,a')|}$
\EndFor
\State \textbf{Return} $a^{\star} \gets \arg\max_{a \in \mathcal{A}} \hat{R}_{u,a}$
\end{algorithmic}
\end{algorithm}

\subsection{Bandit policies}

\subsubsection{TS}

TS is a Bayesian algorithm based on probability matching \citep{thompson1}, where exercises are selected proportionally to their probability of being optimal given the current belief state.
At each round \(t\), the algorithm samples a reward value for each exercise from its posterior distribution and selects the exercise with the highest estimated reward.
This approach naturally balances exploration and exploitation by favoring exercises with high estimated rewards while still exploring uncertain options.

In the standard formulation of TS, the reward is binary, and the posterior distribution over the mean reward of each exercise follows a Beta distribution.
In contrast, in the proposed ERS the reward signal is continuous, representing the learner’s skill gain after engaging with an exercise.
We therefore model rewards as Gaussian-distributed with \emph{unknown} mean and variance.
The conjugate prior for this case is the \emph{Normal–Inverse–Gamma} distribution:
\begin{equation}
P(\mu_a, \sigma_a^2) \sim \mathrm{N}\text{-}\Gamma^{-1}(m_a, \nu_a, \alpha_a, \beta_a),
\end{equation}
where \(m_a\) denotes the prior mean, \(\nu_a > 0\) the prior precision (or pseudo-count), \(\alpha_a > 0\) the prior shape parameter, and \(\beta_a > 0\) the prior scale parameter.
This prior is specified for each exercise $a$, allowing every exercise to maintain its own posterior belief about the expected reward.
Moreover, the prior is conjugate to the Gaussian likelihood with unknown mean and variance, ensuring that the posterior distribution remains in the Normal–Inverse–Gamma family after each update.
A theoretical analysis of this policy has been studied in \citet{honda_theoryGaussian_2014}.

At each round, TS samples from the posterior for every exercise \(a\), selects the exercise with the highest sampled mean, observes the reward signal, and updates the corresponding hyperparameters.
The TS algorithm is formally defined in Algorithm~\ref{alg:ts_nig}.

\begin{algorithm}
\caption{Thompson Sampling with Normal–Inverse–Gamma Prior} \label{alg:ts_nig}
\begin{algorithmic}[1]
\State \textbf{Input:} Exercise set $\mathcal{A}$, hyperparameters $\{m_a, \nu_a, \alpha_a, \beta_a\}_{a \in \mathcal{A}}$
\For{$t = 1, 2, \dots, T$}
  \For{each $a \in \mathcal{A}$}
    \State Sample $\widehat{\sigma_a^2} \sim \mathrm{Inverse\mbox{-}Gamma}(\alpha_a,\beta_a)$
    \State Sample $\widehat{\mu_a} \sim \mathcal{N}\!\big(m_a,\, \widehat{\sigma_a^2}/\nu_a\big)$
  \EndFor
  \State Select $a_t \gets \arg\max_{a \in \mathcal{A}} \widehat{\mu_a}$
  \State Recommend $a_t$ and observe continuous reward $r_{t,a_t}$
  \State \textbf{Update for $a_t:$}
  \State \hspace{4mm} $\tilde m \gets m_{a_t},\ \tilde \nu \gets \nu_{a_t}$
  \State \hspace{4mm} $\nu_{a_t} \gets \tilde \nu + 1$
  \State \hspace{4mm} $m_{a_t} \gets \dfrac{\tilde \nu\, \tilde m + r_{t,a_t}}{\tilde \nu + 1}$
  \State \hspace{4mm} $\alpha_{a_t} \gets \alpha_{a_t} + \tfrac{1}{2}$
  \State \hspace{4mm} $\beta_{a_t} \gets \beta_{a_t} + \dfrac{\tilde \nu\, (\,r_{t,a_t} - \tilde m\,)^2}{2(\tilde \nu + 1)}$
\EndFor
\end{algorithmic}
\end{algorithm}

\subsubsection{LinTS}

LinTS \citep{agrawal2} extends TS to contextual bandits by assuming that the expected reward of each exercise is a linear function of the context vector.
It balances exploration and exploitation by sampling parameter vectors from a Bayesian posterior distribution that captures uncertainty in the estimated parameters.

In LinTS, each exercise \(a\) is associated with an information matrix \(A_a \in \mathbb{R}^{d \times d}\) and a reward vector \(b_a \in \mathbb{R}^d\).
The posterior mean parameter vector \(\omega_{a,t}\) is given by:
\begin{equation}
\omega_{a,t} = A_a^{-1} b_a,
\end{equation}
which is updated iteratively.
The posterior covariance of the parameter estimates is proportional to \(A_a^{-1}\), shrinking as more context–reward pairs are observed for that exercise.

Exploration in LinTS arises naturally from the parameter sampling process.
At each time step \(t\), a parameter vector \(\theta_{a,t}\) is sampled for each exercise \(a\) according to:
\begin{equation}
\theta_{a,t} \sim \mathcal{N}\left(\omega_{a,t}, v^2 A_a^{-1}\right),
\end{equation}
where \(v > 0\) is a fixed exploration scaling factor.
This sampling encourages exploration of exercises with high uncertainty in their reward estimates, while exploitation is driven by the posterior mean \(\omega_{a,t}\).
Over time, as more evidence is gathered and hence \(A_a^{-1}\) shrinks, the algorithm shifts naturally towards exploitation.

In this work, we adopt a Gaussian LinTS variant as formalized in Algorithm~\ref{alg:lints}.
Consistent with the approach of \citet{agrawal2}, the posterior variance scaling factor \( v \) is held fixed rather than estimated adaptively.

\begin{algorithm}
\caption{LinTS Algorithm with Fixed Exploration Scaling Factor \citep{agrawal2}} \label{alg:lints}
\begin{algorithmic}[1]
\State \textbf{Input:} Regularization parameter \( \lambda \), exploration scaling factor \( v \), context space \( \mathcal{X} \), set of exercises \( \mathcal{A} \)
\State Initialize: For each exercise \( a \in \mathcal{A} \),
\[
A_a = \lambda \mathbf{I}_d, \quad b_a = 0^d, \quad \omega_{a,0} = 0^d
\]
\For{each time step \( t = 1, 2, \ldots \)}
  \State Observe context vector \( \mathbf{x}_t \in \mathcal{X} \)
  \For{each exercise \( a \in \mathcal{A} \)}
    \State Sample parameter vector:
    \[
    \theta_{a,t} \sim \mathcal{N}\left(\omega_{a,t}, v^2 A_a^{-1}\right)
    \]
    \State Compute the expected reward:
    \[
    \hat{r}_{a,t} = \mathbf{x}_t^\mathsf{T} \theta_{a,t}
    \]
  \EndFor
  \State Select exercise \( a_t = \arg\max_a \hat{r}_{a,t} \)
  \State Recommend exercise \( a_t \) and observe reward \( r_{a_t,t} \)
  \State Update the selected exercise:
  \[
  A_{a_t} = A_{a_t} + \mathbf{x}_t \mathbf{x}_t^\mathsf{T}
  \]
  \[
  b_{a_t} = b_{a_t} + r_t \mathbf{x}_t
  \]
  \[
  \omega_{a_t, t+1} = A_{a_t}^{-1} b_{a_t}
  \]
\EndFor
\end{algorithmic}
\end{algorithm}

\section{Experimental Setup} \label{sec:exp_settings}

\paragraph{Dataset.}
The dataset used for this experiment is the ASSISTments 2017 dataset \citep{patikorn_assistments_2020}, a large-scale click-stream corpus collected from the web-based ASSISTments tutoring system \citep{heffernan_assistments_2014}, which records middle-school students’ mathematics exercise-solving activities between 2004 and 2006.
The dataset is widely used in learning analytics research, for example in learner performance prediction \citep{soukaina_performance_2024} and knowledge tracing estimation \citep{cully_student_2019, benjamin_student_2024}.
It contains 935{,}638 interaction records from 1{,}708 unique learners across 3{,}162 distinct exercises, with 37.4\% of all attempts answered correctly.

The dataset provides three complementary types of inputs.
First, \textit{clickstream records} capture the raw sequence of learner–system interactions.
Second, \textit{exercise information} describes the exercises themselves, including the associated cognitive skill(s) and exercise type (e.g., multiple-choice, open-response).
Third, \textit{student profiles} encode both background information and behavioral characteristics.
Background attributes include sociodemographic indicators such as gender, the middle school attended, and the academic year in which the system was used.
Behavioral characteristics are derived from historical interaction logs and summarized over the learner’s past activity.
These include:
\begin{enumerate}
  \item Academic proficiency indicators, which include the learner’s average knowledge mastery across all mathematical skills targeted by the system, performance on the Massachusetts Comprehensive Assessment System (MCAS) mathematics test and the learner’s overall correct response rate.
  These academic proficiency indicators capture complementary dimensions of learner ability.
  The MCAS mathematics score provides a stable, externally validated measure of baseline competence, while average knowledge mastery reflects broader conceptual understanding as inferred during system interaction.
  The overall correctness rate indicates how efficiently and accurately learners apply their knowledge.
  \item Affective state indicators that capture internal emotions or psychological states that can influence learning, which include the averaged tendencies towards confusion, frustration, boredom, and engaged concentration.
  \item Disengaged behavior indicators that reflect behavioral patterns that indicate that a learner is not productively engaged, which include the averaged tendencies for carelessness (e.g., slipping an exercise), gaming the system, and disengaging from the learning task.
\end{enumerate}

The affective and disengagement indicators are obtained via a two-stage process: manual labelling through in-class field observations on a representative subsample, followed by the training of automated detectors using supervised machine learning methods to the full dataset \citep{pardos_affective_2014}.
In addition, cognitive measures in the form of BKT estimates of mastery are included in the dataset.
After each exercise attempt, the system recalculates the learner’s probability of mastery for the specific cognitive skill targeted by the exercise, thereby providing a fine-grained, time-varying representation of the learner’s knowledge state.
In this study, reward is defined as \emph{skill gain}, computed from the difference between the BKT mastery estimate after and before the interaction for a specific skill targeted by the exercise.
This continuous-valued signal measures the incremental change in the learner’s estimated mastery due to engaging with the recommended exercise, providing a pedagogically grounded target for optimization.

\begin{figure}[htbp]
    \centering
    \begin{subfigure}{0.48\textwidth}
        \centering
        \includegraphics[width=\linewidth]{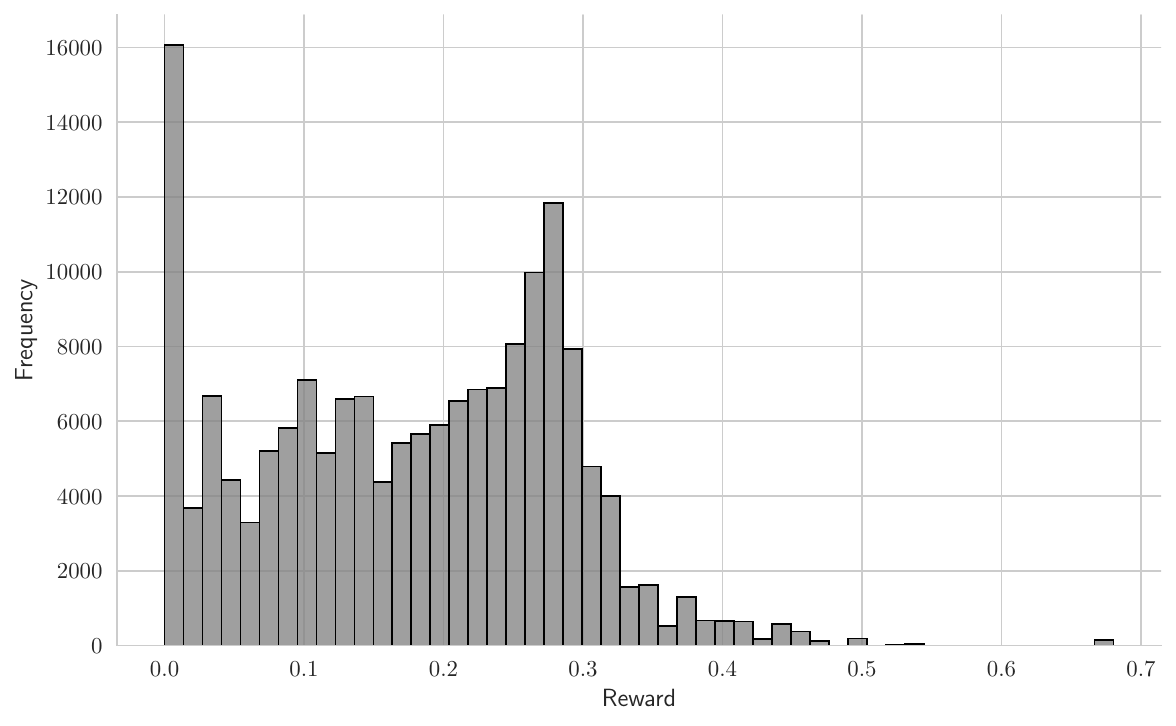}
        \caption{Reward distribution across interactions.}
        \label{fig:reward_distribution}
    \end{subfigure}
    \hfill
    \begin{subfigure}{0.48\textwidth}
        \centering
        \includegraphics[width=\linewidth]{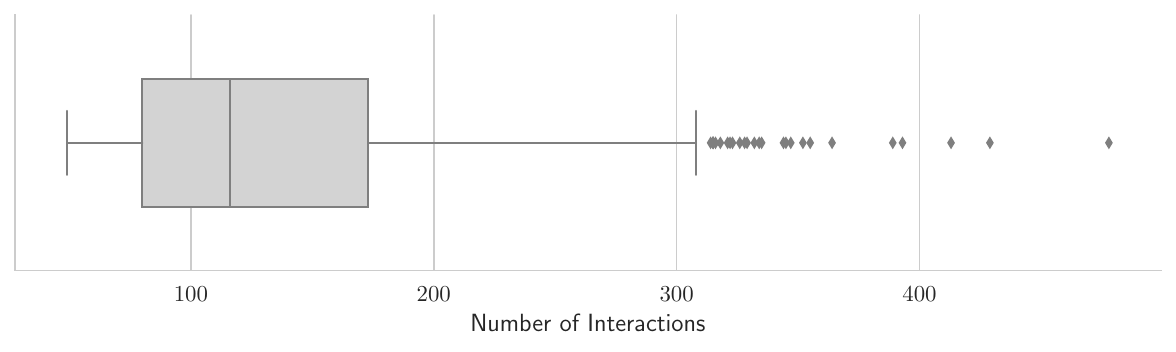}
        \caption{Per-student interaction counts.}
        \label{fig:student_activity}
    \end{subfigure}

    \caption{Overview of dataset characteristics: (a) distribution of skill-gain rewards, and (b) variability in student activity levels.}
    \label{fig:data_overview}
\end{figure}

\paragraph{Data Preprocessing.}
Prior to partitioning the dataset into training, validation, and test subsets, we apply the following preprocessing pipeline.
\begin{enumerate}
    \item \textit{Reward calculation and filtering.}
      Any interaction in which either the pre-interaction mastery estimate or the post-interaction mastery estimate for the skill associated with the attempted exercise is missing is removed.
      Only interactions with strictly positive rewards are retained, focusing the learning process on exercises that have demonstrably advanced a learner’s mastery of the targeted skill and also reducing computation time.
      The empirical distribution of these computed rewards is shown in Figure~\ref{fig:reward_distribution}.
      The pronounced peak around zero reflects the large share of interactions that do not yield measurable improvements in estimated skill gain.
      Moreover, the distribution is positively skewed: most interactions correspond to small to moderate gains, with a long right tail representing relatively larger skill gain improvements.
    \item \textit{Duplicate user–exercise interactions.}
      For any user–exercise pair with multiple recorded interactions, only the most recent chronologically observed attempt is retained.
      This situation often arises on the ASSISTments platform when learners request hints, retry after incorrect responses, or reopen an exercise within the same session.
      Retaining only the final attempt ensures that the post-interaction mastery estimate reflects the learner’s ultimate knowledge state for that exercise, avoiding inflated counts from partial or intermediate states.
    \item \textit{Learner activity threshold.}
      Learners with fewer than $50$ interactions are excluded to preserve sufficient historical data for personalized modelling, as shorter histories produce highly unstable skill gain estimates and provide too little signal for meaningful contextual differentiation.
      The resulting student activity levels, measured as the number of retained interactions per learner, are summarized in Figure~\ref{fig:student_activity}.
    \item \textit{Warm-start enforcement.}
      After the temporal split, any validation or test interactions involving a user or exercise unseen in the training set are removed.
      This ensures that evaluation occurs in a warm-start setting, where all entities at test time have prior representation in the training data, thereby avoiding cold-start scenarios.
\end{enumerate}

A summary of the final preprocessed dataset, including the number of unique users, exercises, skills and total number of interactions is provided in Table~\ref{tab:data_summary}.

\begin{table}[ht]
\centering
\small
\caption{Summary statistics of the preprocessed ASSISTments 2017 dataset.}
\label{tab:data_summary}
\begin{tabularx}{\linewidth}{l r X}
\toprule
\textbf{Statistic} & \textbf{Value} & \textbf{Description} \\
\midrule
Unique users   & 1{,}250   & Number of distinct learners in the dataset. \\
Unique exercises & 2{,}600   & Number of distinct exercises available for recommendation. \\
Interactions   & 167{,}585 & Total number of recorded learner–exercise interactions after preprocessing. \\
Number of skills  & 102     & Number of distinct knowledge concepts, e.g., supplementary angles, Pythagorean theorem. \\
\bottomrule
\end{tabularx}
\end{table}

For the contextual bandit setting, we construct a context vector for each interaction by concatenating user features as listed in Table~\ref{tab:features}.
Categorical variables are one-hot encoded, and continuous features are standardized.

\begin{table}[htp!]
\centering
\caption{Features used to construct the context vector $\mathbf{x}_t$.}
\label{tab:features}
\small
\begin{tabularx}{\linewidth}{l X}
\toprule
\textbf{Feature} & \textbf{Description} \\
\midrule
\multicolumn{2}{l}{\underline{\textbf{Sociodemographic characteristics}}} \\
Academic year & Year(s) during which the learner used the platform (categorical). \\
School & Anonymized middle-school identifier (categorical). \\
Gender & Gender of the learner (categorical). \\[0.2cm]

\multicolumn{2}{l}{\underline{\textbf{Academic proficiency}}} \\
Average knowledge mastery & Average student knowledge level across all skills the learner has attempted. \\
Overall correctness rate & Fraction of correct responses across all attempted exercises. \\
MCAS mathematics score & Standardized MCAS math assessment score. \\[0.2cm]

\multicolumn{2}{l}{\underline{\textbf{Affective state}}} \\
Confusion & Mean predicted probability of confusion over past interactions. \\
Frustration & Mean predicted probability of frustration over past interactions. \\
Boredom & Mean predicted probability of boredom over past interactions. \\
Engaged concentration & Mean predicted probability of being focused/engaged. \\[0.2cm]

\multicolumn{2}{l}{\underline{\textbf{Disengaged behavior}}} \\
Carelessness & Mean predicted probability of careless errors. \\
Gaming the system & Mean predicted probability of exploiting system loopholes. \\
Off-task & Mean predicted probability of disengagement from the learning task. \\
\bottomrule
\end{tabularx}
\end{table}

\paragraph{Data splitting.}
We adopt a temporal user split strategy, a commonly used evaluation approach that splits the historical interactions by percentage based on the interaction timestamps \citep{meng2020_split}.
For each learner, interactions are ordered chronologically, with the first $70\%$ assigned to training, the next $15\%$ to validation, and the final $15\%$ to test.
This preserves the natural temporal sequence of interactions, ensures user overlap across splits, and mirrors real-world online deployment where future learner states are unknown at recommendation time.

\paragraph{Algorithms.}
Both UserCF and ItemCF maintain a user–exercise reward matrix that is updated in buffered batches every 1000 interactions, with pending updates applied at the end of training.
UserCF estimates candidate effectiveness by computing cosine similarity between the target learner and all others.
ItemCF instead relies on similarities between candidate exercises and those previously attempted by the learner.
For the bandit models, TS represents each exercise with a Gaussian reward distribution under a Normal–Inverse–Gamma prior initialized with noninformative hyperparameters, and it updates exercise-specific statistics incrementally after each interaction.
LinTS maintains a separate linear model per exercise with ridge regularization fixed at $\lambda = 1$.
To reduce computational overhead, matrix inversions and parameter estimates are recomputed only every 1000 steps.
Both TS-based models (TS and LinTS) include a short warm-start phase of random exercises to ensure initial coverage.
Finally, in all methods, once a learner has attempted an exercise it is excluded from future recommendations, reflecting realistic tutoring scenarios where repeating the same exercise yields negligible learning gains.

\paragraph{Validation strategy.}
We tune the bandit hyperparameters using grid search on the validation split, using mean instantaneous reward as the evaluation criterion.
LinTS is tuned over different values of the variance-scaling parameter $v$, while TS is tuned over the Normal–Inverse–Gamma prior parameters $(v_0, \alpha_0, \beta_0)$ with a fixed prior mean $m_0=0$.
Table~\ref{tab:hp_grids} summarizes the respective search spaces.
The best-performing configuration for each model is then retrained on the combined training and validation data and evaluated once on the held-out test split, without further adaptation during testing.
The CF baselines have no hyperparameters and are directly trained on the combined dataset before final evaluation, whereas TS-based strategies require hyperparameter tuning, which incurs additional computational cost.

\begin{table}[ht]
\centering
\small
\caption{Hyperparameter grids used for LinTS and TS.}
\label{tab:hp_grids}
\begin{tabularx}{\linewidth}{l l X}
\toprule
\textbf{Model} & \textbf{Hyperparameter} & \textbf{Candidate values} \\
\midrule
LinTS & $v$ & $\{0.001,\,0.01,\,0.05,\,0.1,\,0.25,\,0.5,\,1.0,\,2.0,\,5.0\}$ \\
\midrule
\multirow{4}{*}{TS}
  & $m_0$      & $\{0.0\}$ \\
  & $v_0$      & $\{0.01,\,0.1,\,0.5,\,1.0,\,5.0\}$ \\
  & $\alpha_0$ & $\{0.1,\,1.0,\,2.0\}$ \\
  & $\beta_0$  & $\{0.1,\,1.0,\,2.0\}$ \\
\bottomrule
\end{tabularx}
\end{table}

\section{Results} \label{sec:Results}

All model hyperparameters were selected by maximizing the average instantaneous reward on a validation set.
The best configuration for TS was $
(m_0, v_0, \alpha_0, \beta_0) = (0.0, 0.01, 1.0, 2.0)$,
which corresponds to a neutral prior mean $m_0 = 0$, a low prior precision $v_0 = 0.01$ yielding a diffuse prior over exercise means, and hyperparameters $(\alpha_0, \beta_0) = (1.0, 2.0)$ that maintain posterior uncertainty during the initial learning phase.
For LinTS, the optimal exploration scale was $v = 0.05$, and since $v$ scales the posterior covariance of sampled coefficients, this relatively small value reduces injected sampling noise and thus favors exploitation of the informative learner context once sufficient evidence has been accumulated.

\begin{figure}[htp!]
  \centering
  \includegraphics[width=0.7\linewidth]{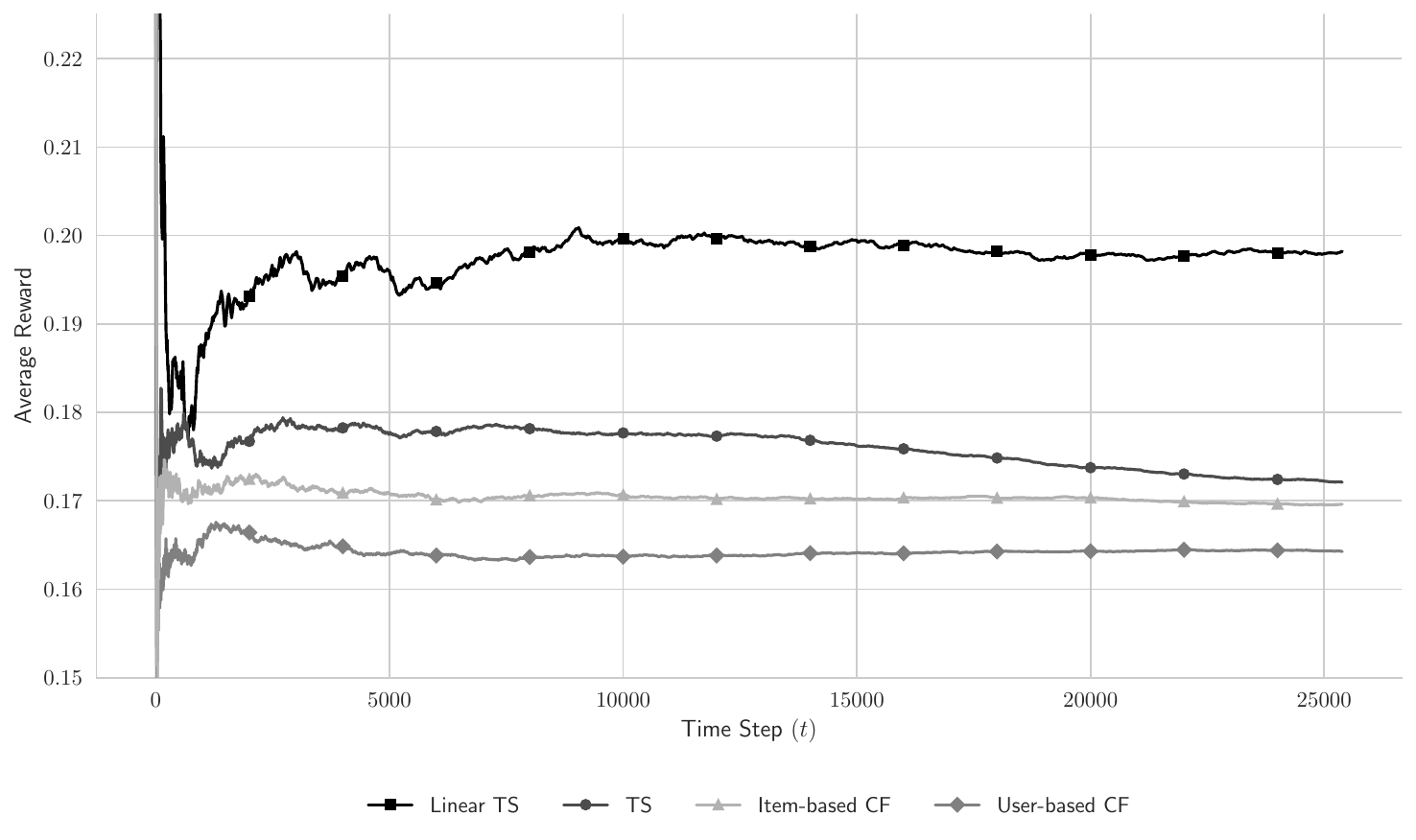}
  \caption{Cumulative average reward on the held-out test set.
   LinTS outperforms all non-contextual baselines, including TS and CF baselines, underscoring the value of contextual modeling in adaptive educational recommendation.}
  \label{fig:test_avg_reward}
\end{figure}

Figure~\ref{fig:test_avg_reward} presents the evolution of cumulative average reward on the held-out test set across all models.
The results show that bandit-based approaches outperform CF baselines.
Both TS variants yield higher rewards than UserCF and ItemCF, confirming that exploration–exploitation strategies can generate more effective recommendations than neighborhood-based heuristics.

Among the tested models, LinTS achieves the highest performance, converging to a final average reward of $0.198$.
This corresponds to a 15.2\% improvement over standard TS ($0.172$), a 16.5\% improvement over ItemCF ($0.170$), and a 20.7\% improvement over UserCF ($0.164$).
While standard TS already performs better than both CF baselines, the contextual extension embodied in LinTS produces substantially larger gains, highlighting the added value of incorporating learner features into the exercise-selection process.

\begin{figure}[t]
    \centering
    \begin{subfigure}{0.48\textwidth}
        \centering
        \includegraphics[width=\linewidth]{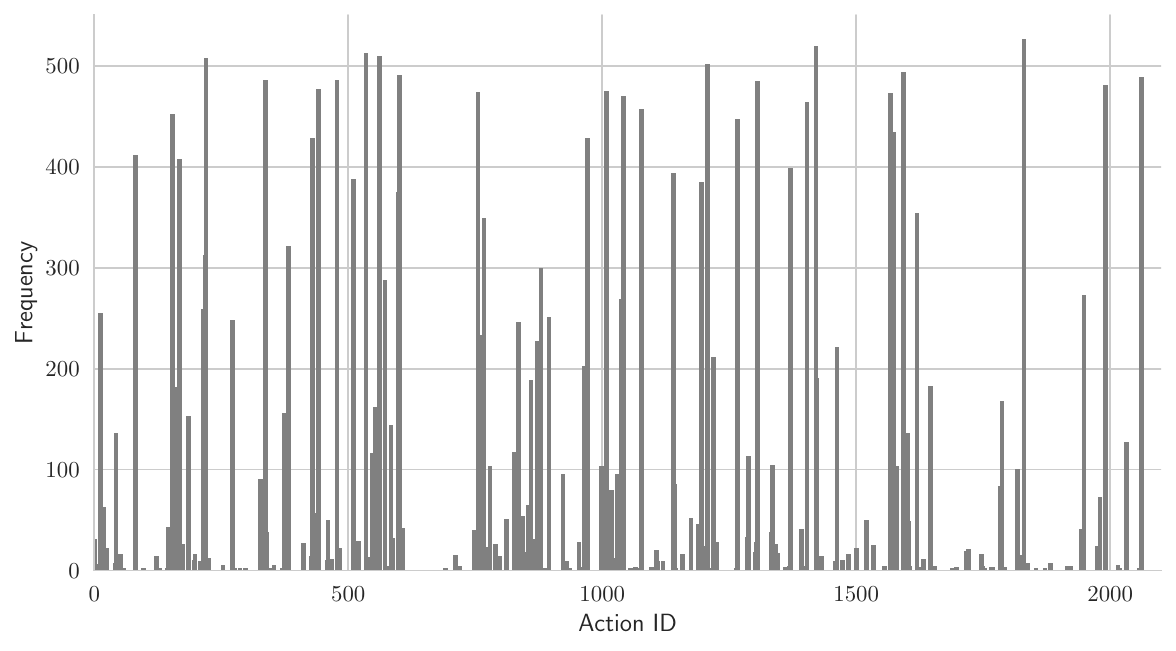}
        \subcaption{ItemCF}
        \label{fig:itemcf_actions}
    \end{subfigure}
    \hfill
    \begin{subfigure}{0.48\textwidth}
        \centering
        \includegraphics[width=\linewidth]{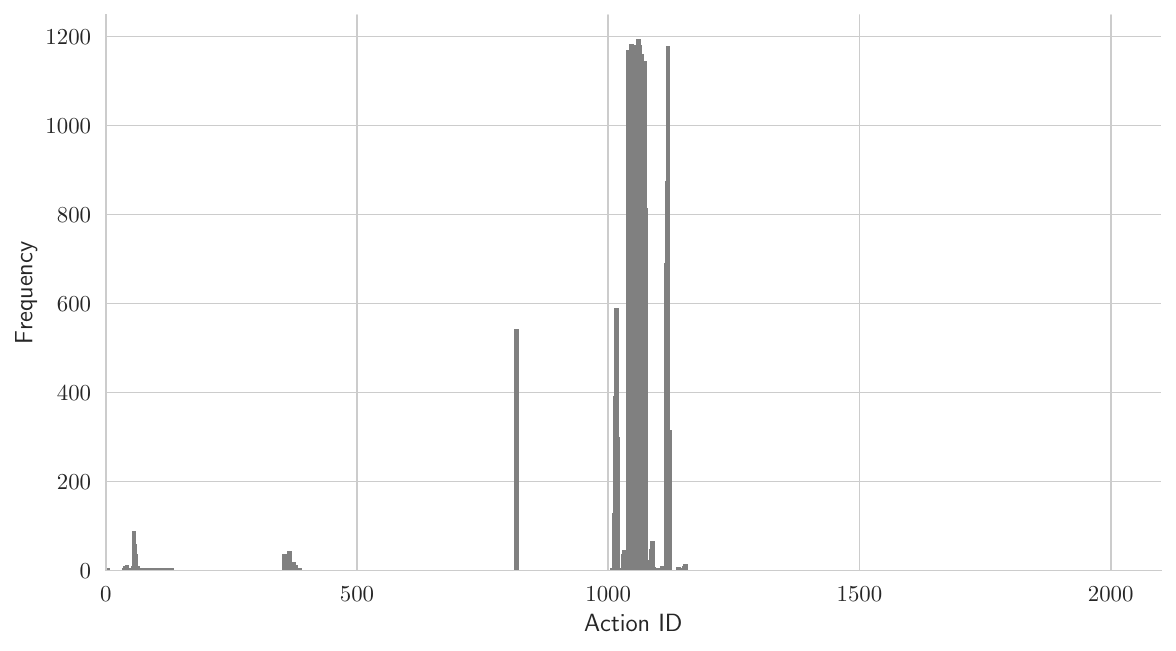}
        \subcaption{UserCF}
        \label{fig:usercf_actions}
    \end{subfigure}

    \vspace{0.3em}

    \begin{subfigure}{0.48\textwidth}
        \centering
        \includegraphics[width=\linewidth]{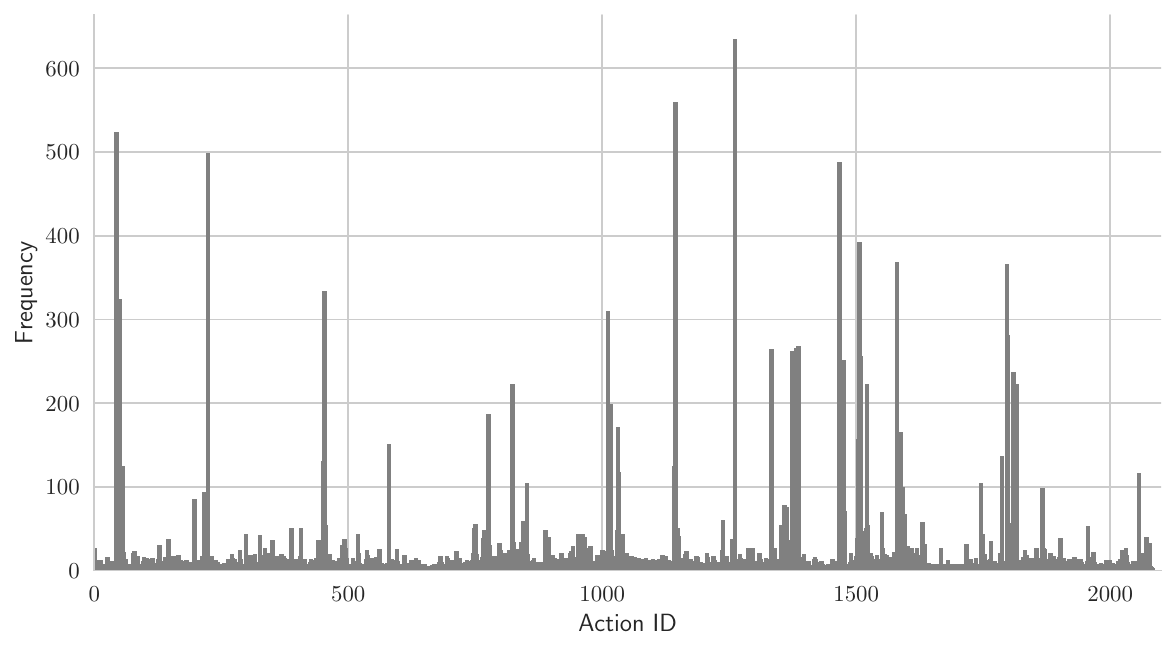}
        \subcaption{TS}
        \label{fig:ts_actions}
    \end{subfigure}
    \hfill
    \begin{subfigure}{0.48\textwidth}
        \centering
        \includegraphics[width=\linewidth]{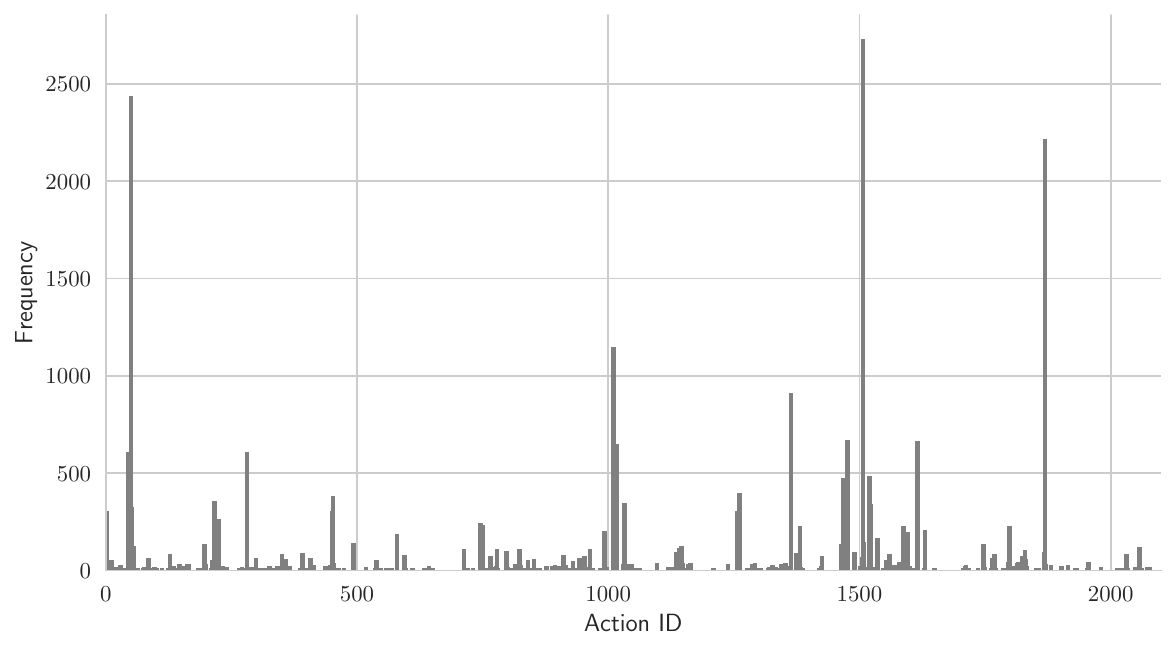}
        \subcaption{LinTS}
        \label{fig:lints_actions}
    \end{subfigure}

    \caption{Exercise selection frequency distributions during testing across the four best-performing agents.
    Contextual modeling (LinTS) concentrates selections on a narrower set of informative exercises, whereas non-contextual strategies spread choices more diffusely across the exercise space.}
    \label{fig:action_freq_comparison}
\end{figure}

Figure~\ref{fig:action_freq_comparison} shows exercise-selection frequency distributions, with the $x$-axis denoting exercise identifiers and the $y$-axis the number of selections during testing.
ItemCF (Figure~\ref{fig:itemcf_actions}) spreads choices widely across the exercise space, reflecting the absence of adaptive prioritization.
UserCF (Figure~\ref{fig:usercf_actions}), by contrast, concentrates almost exclusively on a few exercises, illustrating premature convergence and over-exploitation.
TS (Figure~\ref{fig:ts_actions}) distributes exercises more broadly than UserCF, avoiding the premature lock-in observed in that model.
At the same time, its selections are less diffuse than ItemCF and hence concentrate more on consistently rewarding exercises.
LinTS (Figure~\ref{fig:lints_actions}) goes further by identifying a narrower set of high-value exercises, indicating more effective balancing of exploration and exploitation in the contextual setting.

\begin{figure}[t]
    \centering
    \begin{subfigure}{0.48\textwidth}
        \centering
        \includegraphics[width=\linewidth]{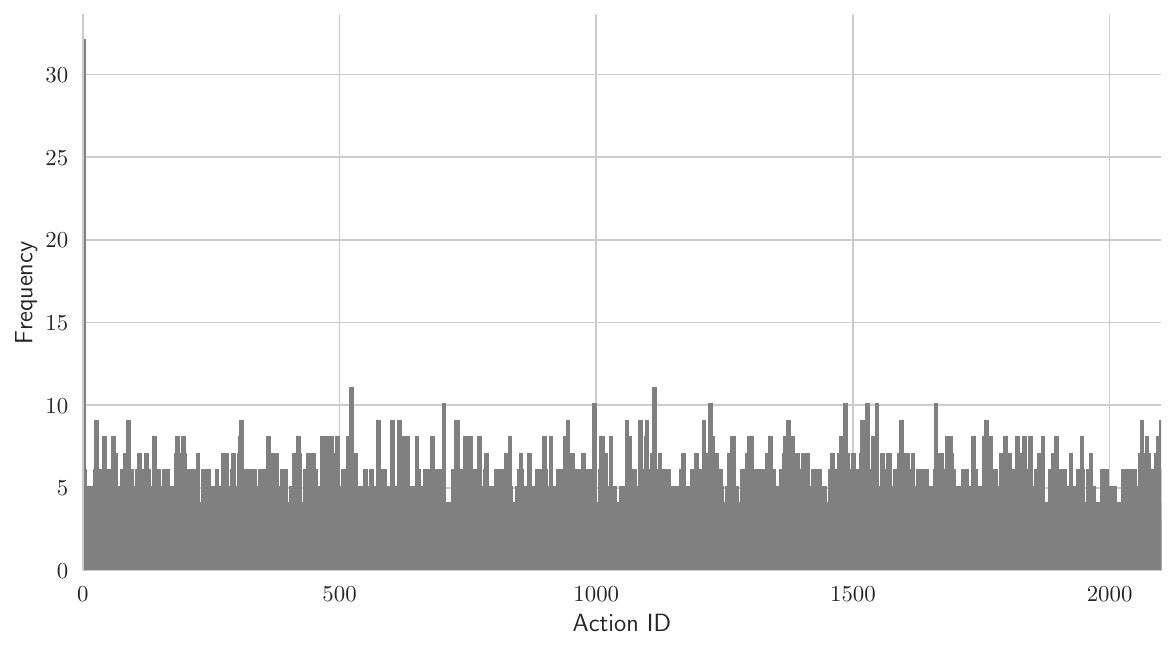}
        \subcaption{Exercise selection frequency after the first 10,000 training rounds.}
        \label{fig:lints_first10k}
    \end{subfigure}
    \hfill
    \begin{subfigure}{0.48\textwidth}
        \centering
        \includegraphics[width=\linewidth]{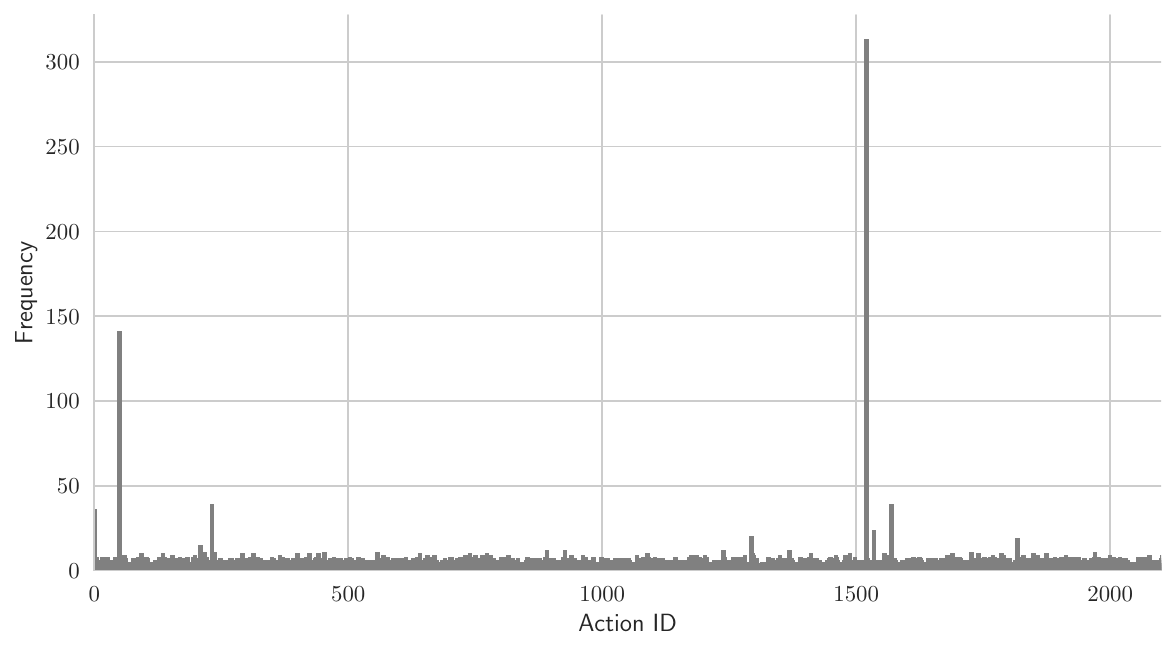}
        \subcaption{Exercise selection frequency for the last 10,000 training rounds.}
        \label{fig:lints_last10k}
    \end{subfigure}
    \caption{Exercise selection frequency distributions for LinTS ($v=0.05$) during training.
     Subfigure~\ref{fig:lints_first10k} shows early exploration behavior, while subfigure~\ref{fig:lints_last10k} illustrates later-stage exploitation dynamics.}
    \label{fig:lints_action_freq}
\end{figure}

To better understand these dynamics, Figure~\ref{fig:lints_action_freq} analyzes LinTS behavior during training.
In the first 10,000 rounds (Figure~\ref{fig:lints_first10k}), the distribution is broad and relatively uniform, reflecting an exploratory phase in which the agent samples widely across the exercise space.
In contrast, during the final 10,000 rounds (Figure~\ref{fig:lints_last10k}), the frequency distribution becomes highly concentrated on a small subset of exercises, indicating focused exploitation of high-value learning opportunities.
These findings highlight that contextual linear modeling not only improves reward performance but also produces qualitatively different exploration–exploitation dynamics, enabling more principled exploration and more focused exploitation of high-reward exercises.

\section{Discussion} \label{sec:Discussion}

From an instructional perspective, the proposed bandit-based ERS offers three key benefits for OR/MS/Analytics courses.
First, many such courses still rely on instructor-centered teaching practices in which all learners progress through the same fixed sequence of exercises predetermined by the instructor.
This structure limits the ability to adjust difficulty, pacing, or feedback to individual needs and often reproduces typical drawbacks of fixed exercise paths: exercises that are too easy may induce disengagement, whereas overly difficult tasks can cause frustration and reduced persistence.
The proposed ERS addresses these limitations by adaptively selecting exercises based on each learner’s evolving skill profile.
For example, the system could be embedded into an introductory optimization or probability course to automatically recommend additional practice on duality or conditional probability for students who struggle, while advancing more prepared students to more difficult exercises.
By personalizing the learning trajectory in this way, LinTS supports scalable learning environments in which instruction becomes responsive rather than prescriptive, enabling individualized practice without requiring instructors to manually construct multiple parallel exercise pathways.
This adaptivity is particularly valuable in large enrollment courses and other digital learning environments where instructors cannot feasibly monitor or tailor learning trajectories for all students.

Second, as shown in Figures~\ref{fig:action_freq_comparison} and~\ref{fig:lints_action_freq}, LinTS concentrates its recommendations on a relatively small set of exercises.
This provides empirically grounded feedback to support course design: the learned policy highlights which exercises consistently generate large skill gains and are therefore strong candidates for in-class discussion, worked examples, or graded assignments.
The same mechanism can underpin an instructor dashboard that identifies which prerequisite skills learners are struggling with, such as matrix operations in linear programming or probability rules in stochastic models, and highlights which exercises produce the strongest gains for specific student subgroups.
Such information supports targeted intervention during classroom hours or tutorial sessions.

Third, learners in OR/MS/Analytics courses often enter with widely varying quantitative competencies in areas such as statistics, linear algebra, and optimization.
As LinTS can condition its recommendations on learner background, it can identify students who struggle with standard exercise sets, for example, because of limited prerequisite skills, and recommend more suitable practice exercises.
This enables instructors to provide differentiated remediation.
Collectively, these insights underscore the potential of contextual bandit–based ERS to support individualized, data-driven instruction and remediation in OR/MS/Analytics courses, including large-scale digital learning environments where traditional personalized feedback is difficult to provide at scale.

\section{Conclusion} \label{sec:Conclusion}

ERS provide a scalable mechanism for supporting active learning in digital OR/MS/Analytics settings, where large and heterogeneous learner populations make individualized guidance difficult to provide manually.
CF remains widely used in ERS, but its reliance on historical similarity patterns, lack of adaptivity, and absence of an exploration mechanism limit its ability to support effective, personalized learning trajectories.
This work introduces a contextual bandit framework based on LinTS, which models exercise effectiveness as a function of learner features and optimizes directly for skill gain.
Experiments on the ASSISTments 2017 dataset show that LinTS outperforms both non-contextual TS and CF baselines, achieving higher average skill gains and exhibiting desirable exploration–exploitation dynamics.
The results highlight several instructional benefits: adaptive sequencing that responds to learners’ evolving skill profiles, data-driven insights into which exercises most effectively promote learning, and the ability to identify students who may require targeted support.

Some limitations of this study must be acknowledged.
The work relies on simplifying assumptions, which may limit the extent to which the findings generalize beyond the present setting.
For example, learners with fewer than 50 interactions were excluded, limiting the applicability of the proposed approach in sparse data settings involving many new or infrequent users.
Future work should incorporate richer contextual signals, explore nonlinear model classes, and consider multi-objective formulations that balance learning progress with other pedagogical goals.
Such extensions would further enhance the applicability of contextual bandits in adaptive learning systems.

\bibliographystyle{informs2014}
\bibliography{references}

%%%%%%%%%%%%%%%%%
\end{document}